# Linear models and linear mixed effects models in R with linguistic applications

Bodo Winter

University of California, Merced, Cognitive and Information Sciences

## Tutorial Part 1: Linear modeling

Linear models and linear mixed models are an impressively powerful and flexible tool for understanding the world. This tutorial is a decidedly conceptual introduction to this class of models. The focus is on **understanding** what these models are doing ... and then we'll spend most of the time **applying** this understanding, using the R statistical programming environment. The idea is to bootstrap your knowledge as quickly as possible so that you can start with your own analyses and then turn to more technical texts if needed. The examples that I draw from come from linguistics and phonetics, but you need not be a linguist to be able to follow this tutorial.

You'll need about 2 hours to complete the full tutorial (maybe a bit more). Each part takes about 1 hour.

---

So, what does the linear model do? Assume you knew nothing about males and females, and you were interested in whether the voice pitch of males and females differs, and if so, by how much.

So you take a bunch of males and a bunch of females, and ask them to say a single word, say "mama", and you measure the respective voice pitches. Your data might look something like this:

| Subject | Sex | Voice.Pitch |
|---|---|---|
| 1 | female | 233 Hz |
| 2 | female | 204 Hz |
| 3 | female | 242 Hz |
| 4 | male | 130 Hz |
| 5 | male | 112 Hz |
| 6 | male | 142 Hz |

"Hz" (Hertz) is a measure of pitch where higher values mean higher pitch.

You might look at this table and say that it's quite obvious that females have higher voice pitch than females. After all, the female values seem to be about 100 Hz above the male ones.

But, in fact, it could be the case that females and males have the same pitch, and you were just unlucky and happened to choose some exceptionally high-pitched females and some exceptionally low-pitched males. Intuitively, the pattern in the table seems pretty straightforward, but we might want a more precise estimate of the difference between males and females, and we might also want an estimate about how likely (or unlikely) that difference in voice pitch could have arisen just because of drawing an unlucky sample.

This is where the linear model comes in. In this case, its task is to give you some values about voice pitch for males and females… as well as some probability value as to how likely those values are.

The basic idea is to express your relationship of interest (in this case, the one between sex and voice pitch) as a simple formula… such as this one:

pitch ~ sex

This reads "pitch predicted by sex" or "pitch as a function of sex". Some people call the thing on the left the "dependent variable" (the thing you measure) and the thing on the right the "independent variable". Others call the thing on the right the "explanatory variable" (this sounds too causal to me) or the "predictor". I'll call it "fixed effect", and this terminology will make sense later on in tutorial 2.

Now, the world isn't perfect. Things aren't quite as deterministic as the above formula suggests. Pitch is not *completely* determined by sex, but also by a bunch of different factors such as language, dialect, personality, age and what not. Even if we measured all of these factors, there would still be other factors influencing pitch that we cannot control for. Perhaps, a subject in your data had a hangover on the morning of the recording (causing the voice to be lower than usual), or the subject was just more nervous on that particular day (causing the voice to be higher). We can never measure and control all of these things. The world is full of stuff that is outside the purview of our little experiment. Hence, let's update our formula to capture the existence of these "random" factors.

>   pitch ~ sex + ε

This "ε" (read "epsilon") is an error term. It stands for all of the things that affect pitch that are not sex, all of the stuff that – from the perspective of our experiment – is random or uncontrollable.

The formula above is a schematic depiction of the linear model that we're going to build. Note that the part of the formula on the right-hand side conceptually divides the world into stuff that you can understand (the "fixed effect" sex) and stuff that you can't understand (the random part "ε"). You could call the former the "structural" or "systematic" part of your model and the latter the "random" or "probabilistic" part of the model.

## Hands-on exercise: Let's start!

O.k., let's move to R, the statistical programming environment that we'll use for the rest of this tutorial[1]. Let's create the dataset that we'll use for our analysis. Type in:

```
pitch = c(233,204,242,130,112,142)
sex = c(rep("female",3),rep("male",3))
```

The first line concatenates our 6 data points from above and saves it in an object that we named `pitch`. The second line repeats the word "female" 3 times and then the word "male" 3 times … and concatenates these 6 words into an object that we named `sex`.

For a better overview, let's combine these two objects into a data frame:

```
my.df = data.frame(sex,pitch)
```

Now we have a data frame object that we named `my.df`, and if you type that, you'll see this:

---

[1] You don't have R? Don't worry, it's free and works on all platforms. You can get it here: http://www.r-project.org/ You might want to read a quick intro to R before you proceed – but even if you don't, you'll be able to follow everything. Just type in everything you see in dark blue.

```
> my.df
    sex pitch
1 female   233
2 female   204
3 female   242
4   male   130
5   male   112
6   male   142
```

O.k., now we'll proceed with the linear model. We take our formula above and feed it into the `lm()` function … except that we omit the "ε" term, because the linear model function doesn't need you to specify this.

```
xmdl = lm(pitch ~ sex, my.df)
```

We modeled pitch as a function of sex, taken from the data frame `my.df` … and we saved this model into an object that we named `xmdl`. To see what the linear model did, we have to "summarize" this object using the function `summary()`:

```
summary(xmdl)
```

If you do this, you should see this:

```
> summary(xmdl)

Call:
lm(formula = pitch ~ sex, data = my.df)

Residuals:
     1       2       3       4       5       6
 6.667 -22.333  15.667   2.000 -16.000  14.000

Coefficients:
            Estimate Std. Error t value Pr(>|t|)
(Intercept)   226.33      10.18  22.224 2.43e-05 ***
sexmale       -98.33      14.40  -6.827  0.00241 **
---
Signif. codes:  0 '***' 0.001 '**' 0.01 '*' 0.05 '.' 0.1 ' ' 1

Residual standard error: 17.64 on 4 degrees of freedom
Multiple R-squared: 0.921,
  Adjusted R-squared: 0.9012
F-statistic: 46.61 on 1 and 4 DF,  p-value: 0.002407
```

Lots of stuff here. First, you're being reminded of the model formula that you entered. Then, the model gives you the residuals (what this is will be discussed later), and the coefficients of the fixed effects (again, explanations follow… bear with me for a moment). Then, the output prints some overall results of the model that you constructed.

We have to work through this output. Let's start with "Multiple R-Squared". This refers to the statistic $R^2$ which is a measure of "variance explained" or if you prefer less causal language, it is a measure of "variance accounted for". $R^2$ values range from 0 to 1. Our $R^2$ is 0.921, which is quite high … you can interpret this as showing that 92.1% of the stuff that's happening in our dataset is "explained" by our model. In this case, because we have only one thing in our model doing the explaining (the fixed effect "sex"), the $R^2$ reflects how much of our data is accounted for by differences between males and females.

In general, you want $R^2$ values to be high, but what is considered a high $R^2$ value depends on your field and on your phenomenon of study. If the system you study is very deterministic, $R^2$ values can be approach 1. But in most of biology and the social sciences, where we study complex or messy systems that are affected by a whole bunch of different phenomena, we frequently deal with much lower $R^2$ values.

The "Adjusted R-squared" value is a slightly different $R^2$ value that not only looks at how much variance is "explained", but also at how many fixed effects you used to do the explaining. So, in the case of our model above, the two values are quite similar to each other, but in some cases the adjusted $R^2_{adj}$ can be much lower if you have a lot of fixed effects (say, you also used age, psychological traits, dialect etc. to predict pitch).

So much for $R^2$. Next line down you see the thing that everybody is crazy for: Your statistical test of "significance". If you've already done research, your eyes will probably immediately jump to the p-value, which in many fields is your ticket for publishing your work. There's a little bit of an obsession with p-values … and even though they are regarded as so important, they are quite often misunderstood! So what exactly does the p-value mean here?

One way to phrase it is to say that *assuming your model is doing nothing, the probability of your data is relatively low* (because the p-value is small in this case). Technically speaking, the p-value is a **conditional probability**, it is a probability *under the condition that the null hypothesis is true*. In this case, the null hypothesis is "sex has no effect on pitch". And, the linear model shows that if this hypothesis is true, then the data would be quite unlikely. This is then interpreted as showing that the alternative hypothesis "sex affects pitch" is more likely and hence that your result is "statistically significant".

Usually, however, you have to distinguish between the significance of the overall model (the p-value at the very bottom of the output), which considers all effects together, from the p-value of individual coefficients (which you find in the coefficients table above the overall significance). We'll talk more about this in a bit.

Then comes the F-value and the degrees of freedom. For an explanation of this, see my tutorial on ANOVAs and the logic behind the F-test (http://bodowinter.com/tutorial/bw_anova_general.pdf). For a general linear model analysis, you probably need this value to report your results. If you wanted to say that your result is "significant", you would have to write something like this:

> "We constructed a linear model of pitch as a function of sex. This model was significant (F(1,4)=46.61, p<0.01). (…)"

Now, let's look at the coefficient table. Here it is again:

```
Coefficients:
            Estimate Std. Error t value Pr(>|t|)
(Intercept)   226.33      10.18  22.224 2.43e-05 ***
sexmale       -98.33      14.40  -6.827  0.00241 **
```

Note that the p-value for the overall model was p=0.002407, which is the same as the p-value on the right-hand side of the coefficients table in the row that starts with "sexmale". This is because your model had only one fixed effect (namely, "sex") and so the significance of the overall model is the same as the significance for this coefficient. If you had multiple fixed effects, then the significance of the overall model and the significance of this coefficient will be different. That is because the significance of the overall model takes all fixed effects (all explanatory variables) into account whereas the coefficients table looks at each fixed effect individually.

But why does it say "sexmale" rather than just "sex", which is how we named our fixed effect? And where did the females go? If you look at the estimate in the row that starts with "(Intercept)", you'll see that the value is 226.33 Hz. This looks like it could be the estimated mean of the female voice pitches. If you type the following…

```
mean(my.df[my.df$sex=="female",]$pitch)
```

… you'll get the mean of female voice pitch values, and you'll see that this value is very similar to the estimate value in the "(Intercept)" column.

Next, note that the estimate for "sexmale" is negative. If you subtract the estimate in the first row from the second, you'll get 128, which is the mean of the male voice pitches (you can verify that by repeating the above command and exchanging "male" for "female").

To sum up, the estimate for "(Intercept)" is the estimate for the female category, and the estimate for "sexmale" is the estimate for the *difference* between the females and the male category. This may seem like a very roundabout way of showing a difference between two categories, so let's unpack this further.

Internally, linear models like to think in lines. So here's a picture of the way the linear model sees your data:

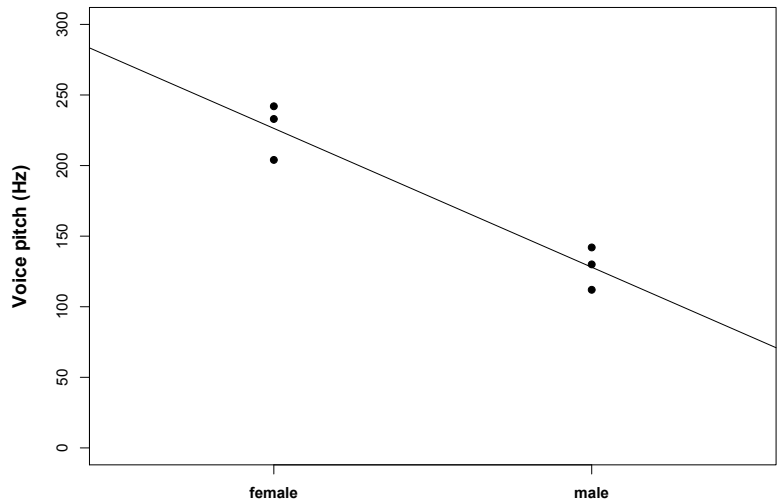

The linear model imagines the difference between males and females as a slope. So, to go "from females to males", you have to go down –98.33 … which is exactly the coefficient that we've seen above. The internal coordinate system looks like this:

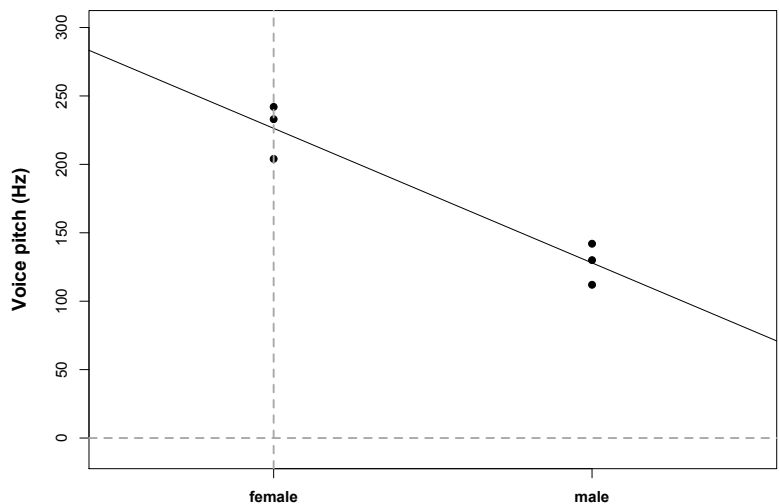

Females are sitting at the x-coordinate zero at the y-intercept (the point where the line crosses the y-axis), and males are sitting at the x-coordinate 1. So now, the output makes a hella more sense to us:

```
Coefficients:
            Estimate Std. Error t value Pr(>|t|)
(Intercept)   226.33      10.18  22.224 2.43e-05 ***
sexmale       -98.33      14.40  -6.827  0.00241 **
```

The females are hidden behind this mysterious "(Intercept)" and the estimate for that intercept is the estimate for female voice pitch! Then, the difference between females and males is expressed as a slope… "going down" by 98.33. The p-values to the right of this table correspond to tests whether each coefficient is "non-zero". Obviously, 226.33 Hz is different from zero, so the intercept is "significant" with a very low p-value. The slope -98.33 is also different from zero (but in the negative direction), and so this is significant as well.

You might ask yourself: Why did the model choose females to be the intercept rather than males? And what is the basis for choosing one reference level over the other? The `lm()` function simply takes whatever comes first in the alphabet! "f" comes before "m", making "females" the intercept at x=0 and "males" the slope of going from 0 to 1.

It might not appear straightforward to you why we can express categorical differences (here, between men and women) as a slope. The reason why this works is because the difference between two categories is exactly correlated with the slope between two categories. The following figures will help you realize this fact. In those pictures, I increased the distance between two categories … and exactly proportional to this increase in distance, the slope increased as well.

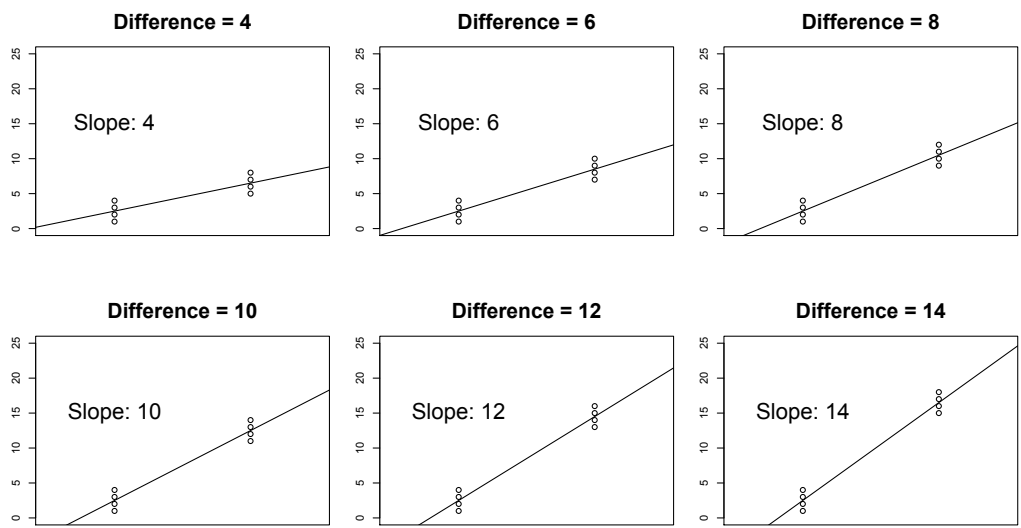

What's the big advantage of thinking of the difference between two categories as a line crossing those two categories? Well, the big advantage is that you can use the same principle for something that is not categorical. So, if you had a continuous factor, say age, you could also fit a line. Everything would work exactly the same. Let's try this out. Say you were now interested in whether age predicts voice pitch. The data might look something like this:

| Subject | Age | Voice.Pitch |
|---------|-----|-------------|
| 1 | 14 | 252 Hz |
| 2 | 23 | 244 Hz |
| 3 | 35 | 240 Hz |
| 4 | 48 | 233 Hz |
| 5 | 52 | 212 Hz |
| 6 | 67 | 204 Hz |

And here's a scatterplot of this data:

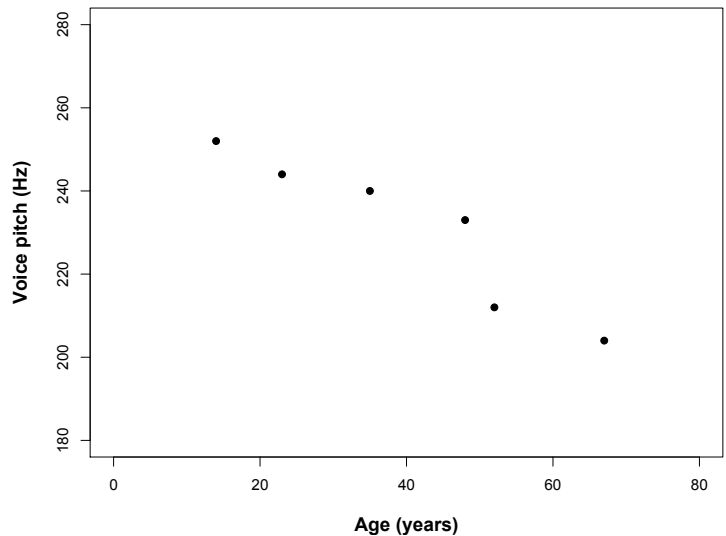

O.k., same thing as before: We express this as a function, where our "fixed effect" is now "age".

$$\text{pitch} \sim \text{age} + \varepsilon$$

Let's construct the data in R and run the model:

```
age = c(14,23,35,48,52,67)
pitch = c(252,244,240,233,212,204)
my.df = data.frame(age,pitch)
xmdl = lm(pitch ~ age, my.df)
summary(xmdl)
```

In the output, let's focus on the coefficients:

```
Coefficients:
            Estimate Std. Error t value Pr(>|t|)
(Intercept) 267.0765     6.8522   38.98 2.59e-06 ***
age          -0.9099     0.1569   -5.80  0.00439 **
```

Again, the significance of the intercept is not very interesting. Remember that the p-value in each row is simply a test of whether the coefficient to the left is significantly different from zero. The intercept (267.0765) here is the predicted pitch value for people with age 0. This intercept doesn't make much sense because people who are not born yet don't really have voice pitch.

What really interests us is "age", which emerges as a significant "predictor" of voice pitch. The way to read the output for age ("-0.9099") is that for every increase of age by 1 you decrease voice pitch by 0.9099 Hertz. Easy-peasy: just go one step to the right in your graph (in your unit of measurement, here: age in years) and one step down (in your unit of measurement, here: voice pitch in Hz).

The scatterplot below neatly summarizes the model: The line represents the mean that the model predicts for people at age 0, 1, 2, 3 etc. This is the line that represents the coefficients of the model. It's worth looking at this picture and comparing it to the coefficients table above. See that the line at x=0 is 267.0765 (our intercept), and the slope is -0.9099.

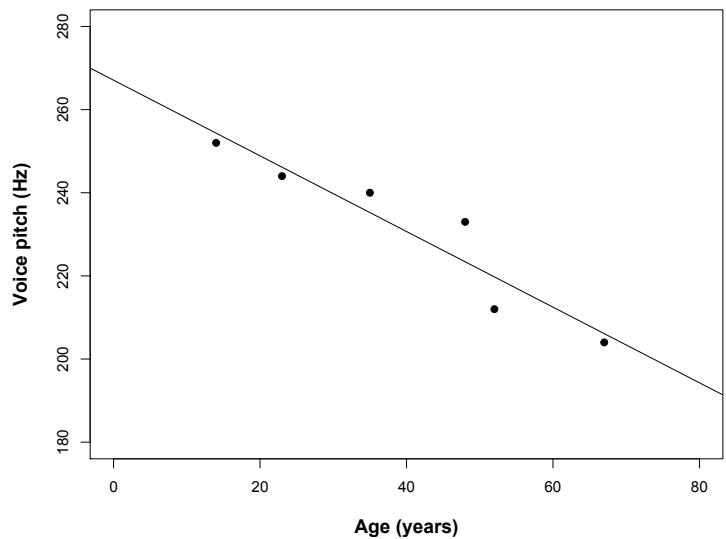

## Meaningful and meaningless intercepts

You might want to remedy the above-discussed situation that the intercept is meaningless. One way of doing this would be to simply subtract the mean age from each age value, as is done below:

```
my.df$age.c = my.df$age – mean(my.df$age)
xmdl = lm(pitch ~ age.c, my.df)
summary(xmdl)
```

Here, we just created a new column "age.c" that is the age variable with the mean subtracted from it. This is the resulting coefficient table from running a linear model analysis of this "centered" data:

```
Coefficients:
            Estimate Std. Error t value Pr(>|t|)
(Intercept) 230.8333     2.8113   82.11 1.32e-07 ***
age.c        -0.9099     0.1569   -5.80  0.00439 **
```

Note that while the estimate has changed from 267.0765 (predicted voice pitch at age 0) to 230.8333 (predicted voice pitch at average age), the slope hasn't changed and neither did the significance associated with the slope or the significance associated with the full model. That is, you haven't messed at all with the nature of your model, you just changed the metric so that the intercept is now the mean voice pitch. So, via centering our variable we made the intercept more meaningful.

## Going on

Both of these examples have been admittedly simple. However, things easily "scale up" to more complicated stuff. Say, you measured two factors ("age" *and* "sex") ... you could put them in the same model. Your formula would then be:

pitch ~ sex + age + ε

Or, you could add dialect as an additional factor:

pitch ~ dialect + sex + age + ε

And so on and so on. The only thing that changes is the following. The p-value at the bottom of the output will be the p-value for the *overall model*. This means that the p-value considers how well all of your fixed effects together help in accounting for variation in pitch. The coefficient output will then have p-values for the individual fixed effects.

This is what people sometimes call "multiple regression", where you model one response variable as a function of multiple predictor variables. The linear model is just another word for multiple regression.

## Assumptions

There's a reason why we call the linear model a _model_. Like any other model, the linear model has ***assumptions*** … and it's important to talk about these assumptions. So here's a whirlwind tour through the conditions that have to be satisfied in order for the linear model to be meaningful:

**(1) Linearity**
It's called "linear model" for a reason! The thing to the left of our simple formula above has to be the result of a linear combination of the things on the right. If it doesn't, the ***residual plot*** will indicate some kind of curve, or it will indicate some other pattern (e.g., two lines if you have categorical binary data).

We haven't talked about residual plots yet, let alone residuals. So, let's do that! Have a look at the picture below, which is a depiction of the age/pitch relationship again:

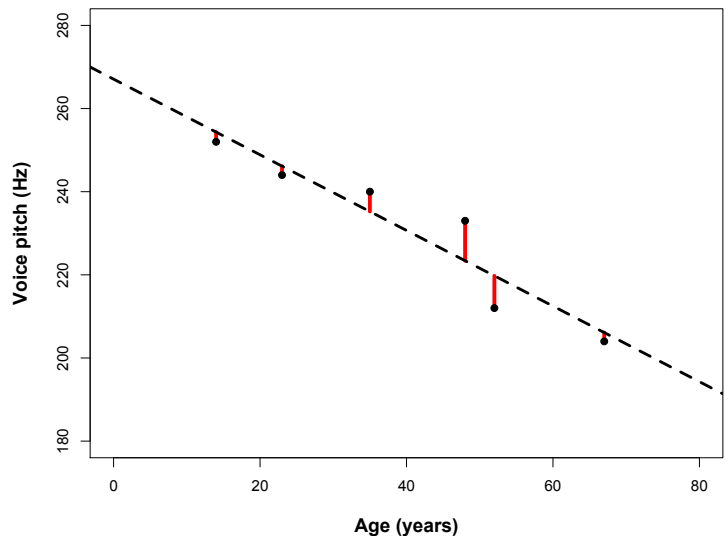

The red lines indicate the residuals, which are the deviations of the observed data points from the predicted values (the so-called "fitted values"). In this case, the residuals are all fairly small … which is because the line that represents the linear model predicts our data very well, i.e., all points are very close to the line.

To get a better view of the residuals, you can take a snapshot of this graph like this…

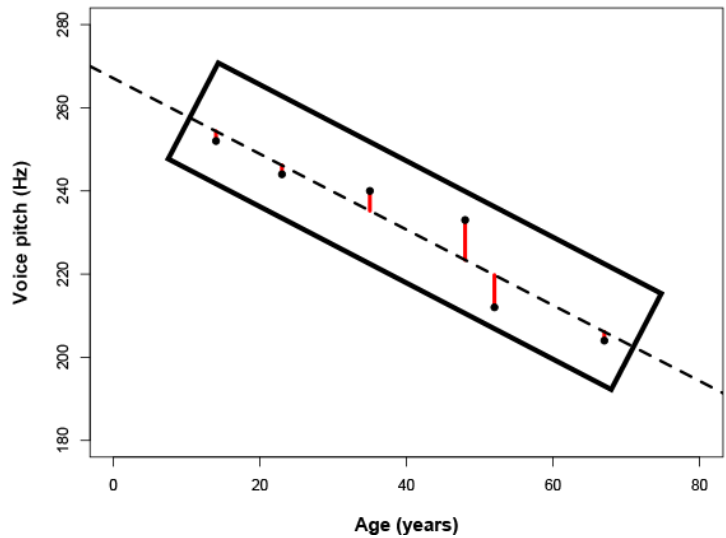

… and rotate it over. So, you make a new plot where the line that the model predicts is now the center line. Like here:

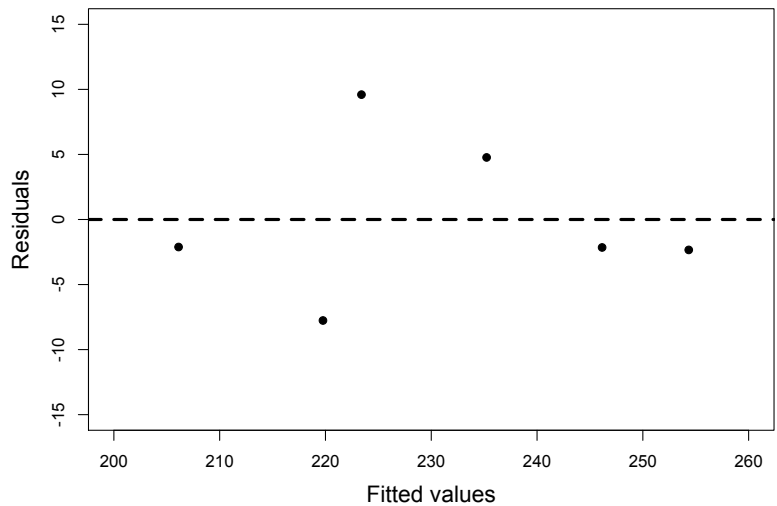

This is a residual plot. The fitted values (the predicted means) are on the horizontal line (at y=0). The residuals are the vertical deviations from this line. This view is just a rotation of the actual data (compare the residual plot with the scatterplot to see this). To construct the residual plot for yourself, simply type:

```
plot(fitted(xmdl),residuals(xmdl))
```
[2]

---

[2] Your plot will have no central line and it will have different scales. It's worth spending some time on tweaking your residual plot and making it pretty… in particular, you should make the plot

In this case… there isn't any obvious pattern in the residuals. If there *were* a nonlinear or curvy pattern, then this would indicate a violation of the linearity assumption. Here's an example of a residual plot that clearly shows a violation of linearity:

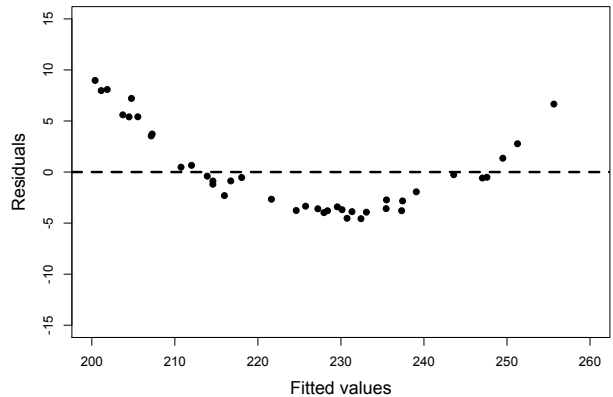

What to do if your residual plot indicates nonlinearity? There's several options:

- You might miss an important fixed effect that interacts with whatever fixed effects you already have in your model. Potentially the pattern in the residual plot goes away if this fixed effect is added.
- Another (commonly chosen) option is to perform a nonlinear transformation of your response, e.g., by taking the log-transform.
- You can also perform a nonlinear transformation of your fixed effects. So, if age were somehow related to pitch in a U-shaped way (perhaps, if very young people had high pitch and very old people had high pitch, too, with intermediate ages having a "dip" in pitch), then you could add age and age$^2$ (age-squared) as predictors.
- Finally, if you're seeing stripes in your residual plot, then you're most likely dealing with some kind of categorical data – and you would need to turn to a somewhat different class of models, such as logistic models.

---

so that there's more space around the margins. This will make any patterns easier to see. Have a look at some R graphic tutorials for this.

**(2) Absence of collinearity**
When two fixed effects (two predictors) are correlated with each other, they are said to be *collinear*. Say, you were interested in how average talking speed affects intelligence ratings (i.e., people who talk more quickly are rated to be more intelligent)…

>  intelligence ratings ~ talking speed

… and you measured several different indicators of talking speed, for example, you syllables per seconds, words per seconds and sentences per seconds. These different measures are going to be correlated with each other because if you speak more quickly, then you say more syllables, words and sentences in a given amount of time. If you'd use all of these correlated predictors to predict intelligence ratings within the same model, you are likely going to run into a collinearity problem.

If there's collinearity, the interpretation of the model becomes unstable: Depending on which one of correlated predictors is in the model, the fixed effects become significant or cease to be significant. And, the significance of these correlated or collinear fixed effects is not easily interpretable, because they might steal each other's "explanatory power" (that's a very coarse way of saying what's actually going on, but you get the idea).

Intuitively, this makes a lot of sense: If multiple predictors are very similar to each other, then it becomes very difficult to decide what, in fact, is playing a big role.

How to get rid of collinearity? Well first of all, you might pre-empt the problem in the design stage of your study and focus on a few fixed effects that you know are not correlated with each other. If you didn't do this and you have several multiple measures to choose from at the analysis stage of your study (e.g., three different ways of measuring "talking speed"), think about which one is the most meaningful and drop the others (be careful here: don't base this dropping decision on the "significance"). Finally, you might want to consider dimension-reduction techniques such as Principal Component Analysis. These can transform several correlated variables into a smaller set of variables which you can then use as new fixed effects.

**(3) Homoskedasticity … or "absence of heteroskedasticity"**
Being able to pronounce "heteroskedasticity" several times in a row in quick succession will make you a star at your next cocktail party, so go ahead and rehearse pronouncing them now!

Jokes aside, homoskedasticity is an extremely important assumption. It says that the variability of your data should be approximately equal across the range of your predicted values. If homoscedasticity is violated, you end up with heteroskedasticity, or, in other words, a problem with unequal variances.

For the homoscedasticity assumption to be met, the **_residuals_** of your model need to roughly have a similar amount of deviation from your predicted values. Again, we can check this by looking at a residual plot. Here's the one for the age/pitch data again:

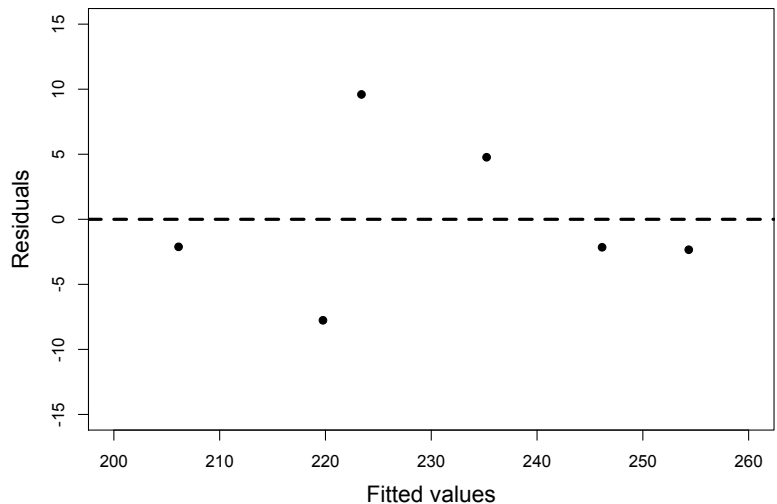

There's not really that many data points to tell whether this is really homoscedastic. In this case, I would conclude that there's not enough data to safely determine whether there is or isn't heteroskedasticity. Usually, I would construct models for much larger data sets anyway.

So, here's a plot that gives you an idea of how a "good" residual plot looks with more data:

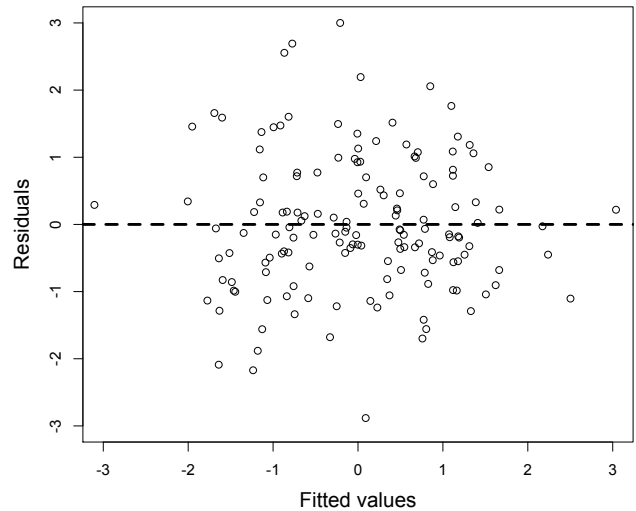

And another one:

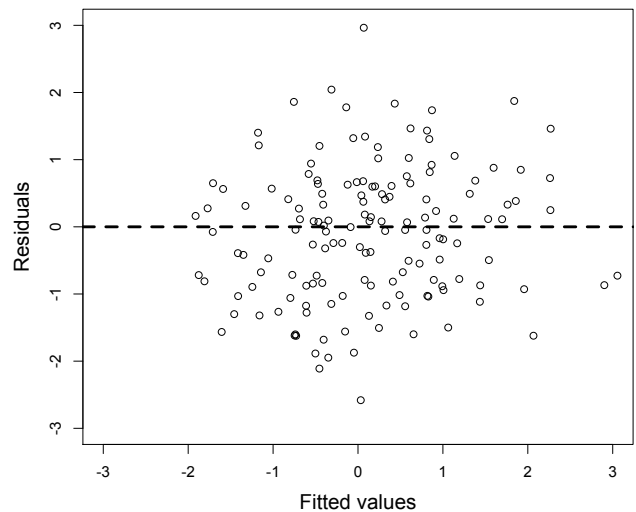

A good residual plot essentially looks blob-like. It's a good idea to generate some random data to see how a plot with roughly equal variances looks like. You can do so using the following command line:

```
plot(rnorm(100),rnorm(100))
```

This creates two sets of 100 normally distributed random numbers with a mean of 0 and a standard deviation of 1. If you type this in multiple times to create multiple plots, you can get a feel of how a "normal" residual plot should look like.

The next residual plot shows obvious heteroskedasticity:

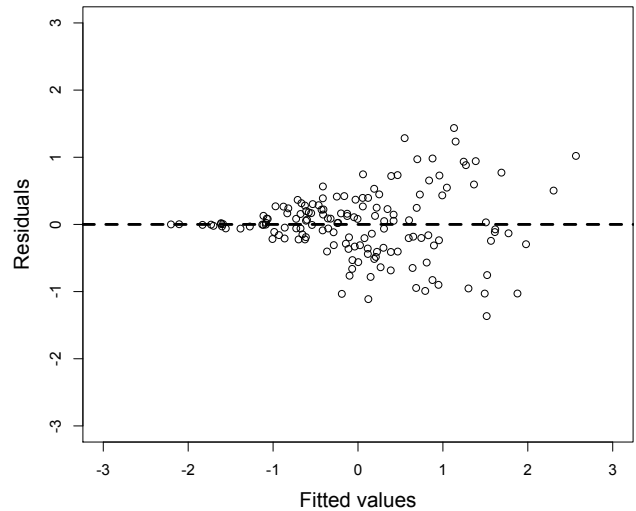

In this plot, higher fitted values have larger residuals ... indicating that the model is more "off" with larger predicted means. So, the variability is not homoscedastic: it's smaller in the lower range and larger in the higher range.

What to do? Again, transforming your data often helps. Consider a log-transform here as well.

**(4) Normality of residuals**
The normality of residuals assumption is the one that is least important. Interestingly, many people seem to think it is the most important one, but it turns out that linear models are relatively robust against violations of the assumptions of normality. Researchers differ with respect to how much weight they put onto checking this assumption. For example, Gellman and Hill (2007), a famous book on linear models and mixed models, do not even recommend diagnostics of the normality assumption (ibid. 46).

If you wanted to test the assumption, how would you do this? Either you make a histogram of the residuals of your model, using...

```
hist(residuals(xmdl))
```

... or a Q-Q plot ...

```
qqnorm(residuals(xmdl))
```

Here's a residual plot and a Q-Q plot of the same data next to each other.

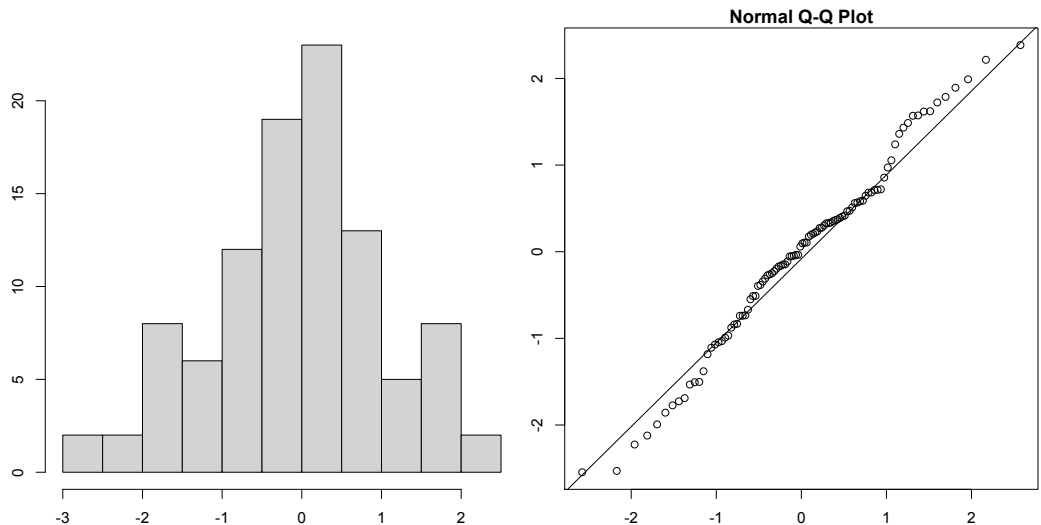

These look good. The histogram is relatively bell-shaped and the Q-Q plot indicates that the data falls on a straight line (which means that it's similar to a normal distribution). Here, we would conclude that there are no obvious violations of the normality assumption.

**(5) Absence of influential data points**
Some people wouldn't call "the absence of influential data points" an *assumption* of the model. However, influential data points can drastically change the interpretation of your results, and similar to collinearity, it can lead to instable results.

How to check? Here's a useful R function, `dfbeta()`, that you can use on a model object like our `xmdl` from above.

```
> dfbeta(xmdl)
  (Intercept)         age
1  -3.3645662  0.06437573
2  -1.6119656  0.02736278
3   1.5481303 -0.01456709
4  -0.0259835  0.05092767
5   0.8707699 -0.06479736
6   1.8551808 -0.06622744
```

For each coefficient of your model (including the intercept), the function gives you the so-called DFbeta values. These are the values with which the coefficients have to be adjusted if a particular data point is excluded (sometimes called "leave-one-out diagnostics"). More concretely, let's look at the age column in the data frame above. The first row means that the coefficient for age (which, if you remember, was -0.9099) has to be adjusted by 0.06437573 if data point 1 is excluded. That means that the coefficient of the model without the data point would be -0.9742451 (which is -0.9099 minus 0.06437573… if the slope is negative, DFbeta values are subtracted, if it's positive, they are added).

There's a little bit of room for interpretation in what constitutes a large or a small DFbeta value. One thing you can say for sure: Any value that changes the sign of the slope is *definitely* an influential point that warrants special attention… because excluding that point would change the interpretation of your results. What I then do is to eyeball the DFbetas and look for values that are different by half of the absolute value of the slope. Say, my slope would be 2 … then a DFbeta value of 1 or -1 would be alarming to me. If it's a slope of -4, a DFbeta value of 2 or -2 would be alarming to me.

How to proceed if you have influential data points? Well, it's definitely not legit to simply exclude those points and report only the results on the reduced data set. A better approach would be to run the analysis *with* the influential points and then again *without* the influential points … then you can report both analyses and state whether the interpretation of the results does or doesn't change. The only case when it is o.k. to exclude influential points is when there's an obvious error with them, so for example, a value that doesn't make sense (e.g., negative age, negative height) or a value that obviously is the result due to a technical error in the data acquisition stage (e.g., voice pitch values of 0). Influence diagnostics allow you to spot those points and you can then go back to the original data and see what went wrong[3].

**(6) Independence !!!!!!**
The independence assumption is by far the most important assumption of all statistical tests. In the linear model analyses that we did so far, each data point came from a different subject. To remind you, here's our two data sets that we worked on:

| **Study 1** | | | **Study 2** | | |
|---|---|---|---|---|---|
| **Subject** | **Sex** | **Voice.Pitch** | **Subject** | **Age** | **Voice.Pitch** |
| 1 | female | 233 Hz | 1 | 14 | 252 Hz |
| 2 | female | 204 Hz | 2 | 23 | 244 Hz |
| 3 | female | 242 Hz | 3 | 35 | 240 Hz |
| 4 | male | 130 Hz | 4 | 48 | 233 Hz |
| 5 | male | 112 Hz | 5 | 52 | 212 Hz |
| 6 | male | 142 Hz | 6 | 67 | 204 Hz |

We were able to run the linear model on this data the way we did only because each row in this dataset comes from a different subject. If you elicit multiple responses from each subject, then those responses that come from the same subject cannot be regarded as independent from each other.

---

[3] For an interesting back-and-forth on a particular example of how much influence diagnostics and extreme values can change the interpretation of a study, have a look at the delightful episode of academic banter between Ullrich and Schlüter (2011) and Brandt (2011).

So, what exactly is independence? The ideal case is a coin flip or the roll of a die: Each coin flip and each roll of a die is absolutely independent from the outcome of the preceding coin flips or die rolls. The same should hold for your data points when you run a linear model analysis. So, the data points should come from different subjects. And each subject should only contribute one data point.

When you violate the independence assumption, all hell breaks loose. The other assumptions that we mentioned above are important as well, but the independence assumption is by far the most important one. Violating independence may greatly inflate your chance of finding a spurious result and it results in a p-value that is completely meaningless. Unfortunately, violations of the independence assumption are quite frequent in many branches of science – so much in fact, that there's a whole literature associated with this violation, starting from Hurlbert (1984) for ecology, Freeberg and Lucas (2009) for psychology, Lazic (2010) for neuroscience and my own small paper for phonetics/speech science (Winter, 2011).

How can you guarantee independence? Well, independence is a question of the experimental design in relation to the statistical test that you use. Design and statistical analyses are closely intertwined and you can make sure that you meet the independence assumption by only collecting one data point per subject.

Now, a lot of the times, we want to collect more data per subject, such as in repeated measures designs. If you end up a data set that has non-independencies in it, you need to **resolve these non-independencies** at the analysis stage. This is where **mixed models** come in handy… and this is where we'll switch to the second tutorial.

# Part 2: A very basic tutorial for performing linear mixed effects analyses

This part serves as a quick boot camp to jump-start your own analyses with linear mixed effects models. This text is different from other introductions by being decidedly *conceptual*; I will focus on *why* you want to use mixed models and *how* you should use them. While many introductions to this topic can be very daunting to readers who lake the appropriate statistical background, this text is going to be a softer kind of introduction… so, don't panic!

Part 2 will take you about 1 hour (possibly a bit more).

## Introduction: Fixed and random effects

In tutorial 1, we talked about how we could use the linear model to express the relationships in our data in terms of a function. In one example, we modeled pitch as a function of age.

pitch ~ age + ε

We called "age" a fixed effect, and ε was our "error term" to represent the deviations from our predictions due to "random" factors that we cannot control experimentally. You could call this part the "probabilistic" or "stochastic" part of the model. Now, we'll unpack this "ε" and add complexity to it. That is, we change the random aspect of our model, essentially leaving the systematic part unchanged. In mixed models, everything in the "systematic" part of your model works just like with linear models in tutorial 1.

In one of my studies, we have been interested in the relationship between pitch and politeness (Winter & Grawunder, 2012). So, essentially we're aiming for a relationship that looks like something like this:

pitch ~ politeness + ε

In our study, politeness was treated as a categorical factor with two levels… a formal register and an informal register. On top of that, we also had an additional fixed effect, sex, and so our formula looks more like this:

pitch ~ politeness + sex + ε

So far so good. Now things get a little more complicated. Our design was so that we took multiple measures per subject. That is, each subject gave multiple polite responses and multiple informal responses. If we go back to the discussion of the assumptions of the linear model in tutorial 1, we can immediately see that this would violate the independence assumption: Multiple responses from the same subject cannot be regarded as independent from each other. Every person has a slightly different voice pitch, and this is going to be an idiosyncratic factor that affects all responses from the same subject, thus rendering these different responses inter-dependent rather than independent.

The way we're going to deal with this situation is to add a ***random effect*** for subject. This allows us to resolve this non-independence by assuming a different "baseline" pitch value for each subject. So, subject 1 may have a mean voice pitch of 233 Hz across different utterances, and subject 2 may have a mean voice pitch of 210 Hz per subject. Here's a visual depiction of how this looks like:

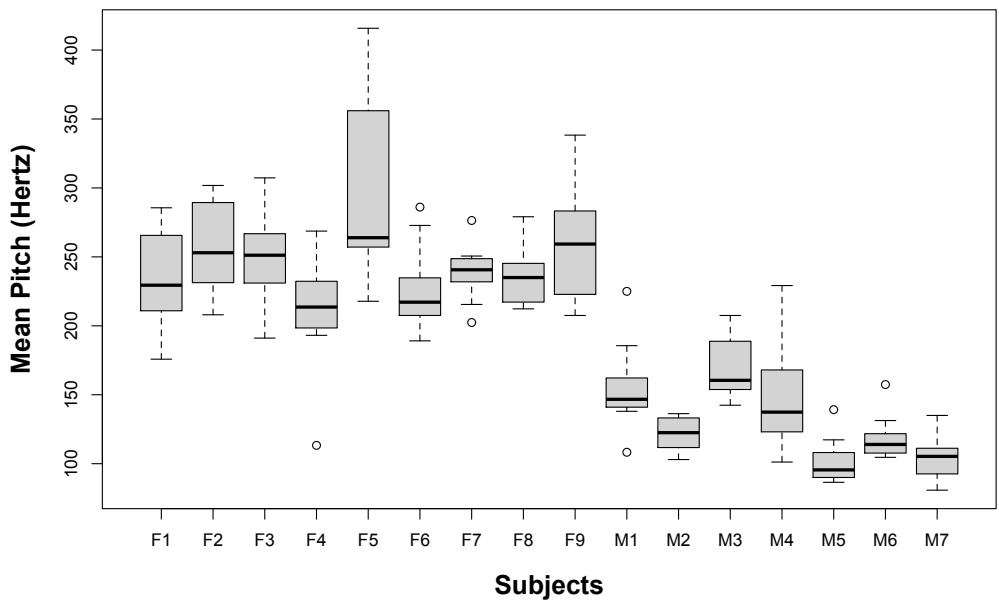

Subjects F1 to F9 are female subjects. Subjects M1 to M7 are male subjects. You immediately see that males have lower voices than females (as is to be expected). But on top of that, within the male and the female groups, you see lots of individual variation, with some people having relatively higher values for their sex and others having relatively lower values.

We can model these individual differences by assuming different ***random intercepts*** for each subject. That is, each subject is assigned a different intercept value, and the mixed model estimates these intercepts for you.

Now you begin to see why the mixed model is called a "mixed" model. The linear models that we considered so far have been "fixed-effects-only" models that had one or more fixed effects and a general error term "ε". With the linear model, we essentially divided the world into things that we somehow understand or that are somehow systematic (the fixed effects, or the explanatory variables); and things that we cannot control for or that we don't understand (ε). But crucially, this latter part, the unsystematic part of the model, did not have any interesting structure. We simply had a general across-the-board error term.

In the mixed model, we add one or more random effects to our fixed effects. These random effects essentially give structure to the error term "ε". In the case of our model here, we add a random effect for "subject", and this characterizes idiosyncratic variation that is due to individual differences.

The mixture of fixed and random effects is what makes the mixed model a mixed model.

Our updated formula looks like this:

> pitch ~ politeness + sex + (1|subject) + ε

"(1|subject)" looks a little enigmatic. I'm already using the R-typical notation format here. What this is saying is "assume an intercept that's different for each subject" … and "1" stands for the intercept here. You can think of this formula as telling your model that it should expect that there's going to be multiple responses per subject, and these responses will depend on each subject's baseline level. This effectively resolves the non-independence that stems from having multiple responses by the same subject.

Note that the formula still contains a general error term "ε". This is necessary because even if we accounted for individual by-subject variation, there's still going to be "random" differences between different utterances from the same subject.

O.k., so far so good. But we're not done yet. In the design that we used in Winter and Grawunder (2012), there's an additional source of non-independence that needs to be accounted for: We had different items. One item, for example, was an "asking for a favor" scenario. Here, subjects had to imagine asking a professor for a favor (polite condition), or asking a peer for a favor (informal condition). Another item was an "excusing for coming too late" scenario, which was similarly divided between polite and informal. In total, there were 7 such different items.

Similar to the case of by-subject variation, we also expect by-item variation. For example, there might be something special about "excusing for coming too late" which leads to overall higher pitch (maybe because it's more embarrassing than asking for a favor), regardless of the influence of politeness. And whatever it is that makes one item different from another, the responses of the different subjects in our experiment might similarly be affected by this random factor that is due to item-specific idiosyncrasies. That is, if "excusing for coming to late" leads to high pitch (for whatever reason), it's going to do so for subject 1, subject 2, subject 3 and so on. Thus, the different responses to one item cannot be regarded as independent, or, in other words, there's something similar to multiple responses to the same item – even if they come from different people. Again, if we did not account for these interdependencies, we would violate the independence assumption.

Here's a visual representation of the by-item variability:

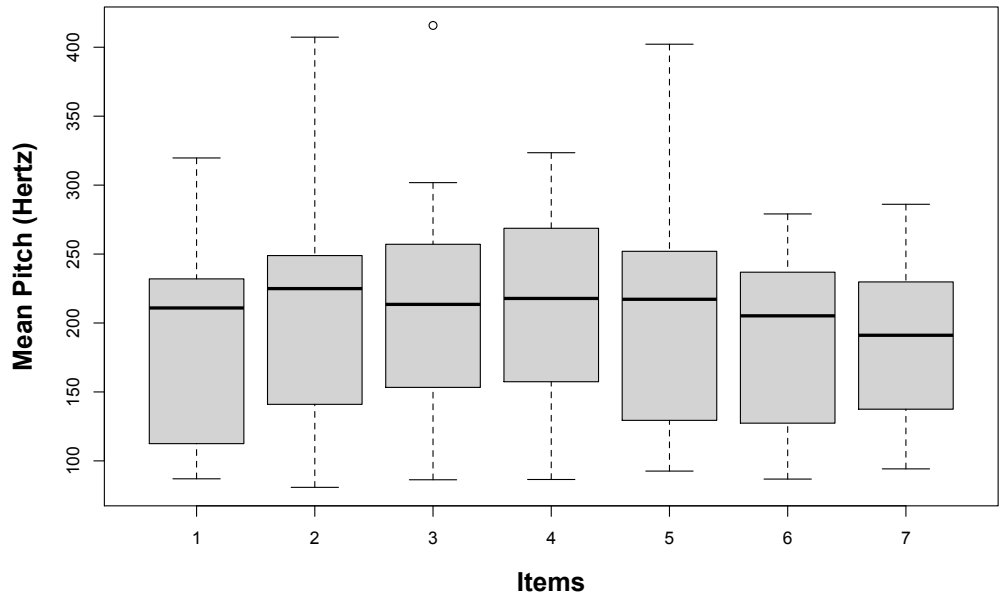

The variation between items isn't as big as the variation between subjects – but there are still noticeable differences, and we better account for them in our model!

We do this by adding an additional random effect:

    pitch ~ politeness + sex + (1|subject) + (1|item) + ε

So, on top of different intercepts for different subjects, we now also have different intercepts for different items. We now "resolved" those non-independencies (our model knows that there are multiple responses per subject and per item), and we accounted for by-subject and by-item variation in overall pitch levels.

Note the efficiency and elegance of this model. Before, people used to do a lot of averaging. For example, in psycholinguistics, people would average over items for a subjects-analysis (each data point comes from one subject, assuring independence), and then they would also average over subjects for an items-analysis (each data point comes from one item). There's a whole literature on the advantages and disadvantages of this approach (Clark, 1973; Forster & Dickinson, 1976; Wike & Church, 1976; Raaijmakers, Schrijnemakers, & Gremmen, 1999; Raaijmakers, 2003; Locker, Hoffman, & Bovaird, 2007; Baayen, Davidson, & Bates, 2008; Barr, Levy, Scheepers, & Tilly, 2013).

The upshot is: while traditional analyses that do averaging are in principle legit, mixed models give you much more flexibility … and they take the full data into account. If you do a subjects-analysis (averaging over items), you're essentially *disregarding* by-item variation. Conversely, in the items-analysis, you're disregarding by-subject variation. Mixed models account for both sources of variation *in a single model*. Neat, init?

Let's move on to R and apply our current understanding of the linear mixed effects model!!

**Mixed models in R**

For a start, we need to install the R package *lme4* (Bates, Maechler & Bolker, 2012). While being connected to the internet, open R and type in:

```
install.packages("lme4")
```

Select a server close to you. After installation, load the *lme4* package into R with the following command:

```
library(lme4)
```

Now, you have the function `lmer()` available to you, which is the mixed model equivalent of the function `lm()` in tutorial 1. This function is going to construct mixed models for us.

But first, we need some data! I put a shortened version of the dataset that we used for Winter and Grawunder (2012) onto my server. You can load it into R the following way:

```
politeness= read.csv("http://www.bodowinter.com/tutorial/politeness_data.csv")
```

Or you can download it by hand…

http://www.bodowinter.com/tutorial/politeness_data.csv

…and load it into R the following way:

```
politeness = read.csv(file.choose( ))
```

Now, you have a data frame called `politeness` in your R environment. You can familiarize yourself with the data by using `head()`, `tail()`, `summary()`, `str()`, `colnames()`… or whatever commands you commonly use to get an overview of a dataset. Also, it is always good to check for missing values:

```
        which(is.na(politeness)==T)
```

Apparently, there is a missing value in row 263. This is important to know but fortunately, a few missing values provide no problems for our mixed model analyses.

The difference in politeness level is represented in the column called "attitude". In that column, "pol" stands for polite and "inf" for informal. Sex is represented as "F" and "M" in the column "gender". The dependent measure is "frequency", which is the voice pitch measured in Hertz (Hz). To remind you, higher values mean higher pitch.

The interesting random effects for us are in the column "subject" and "scenario", the latter being the name of the item column (remember the different scenarios like "asking for a favor"?).

Let's look at the relationship between politeness and pitch by means of a boxplot:

```
        boxplot(frequency ~ attitude*gender,
            col=c("white","lightgray"),politeness)
```

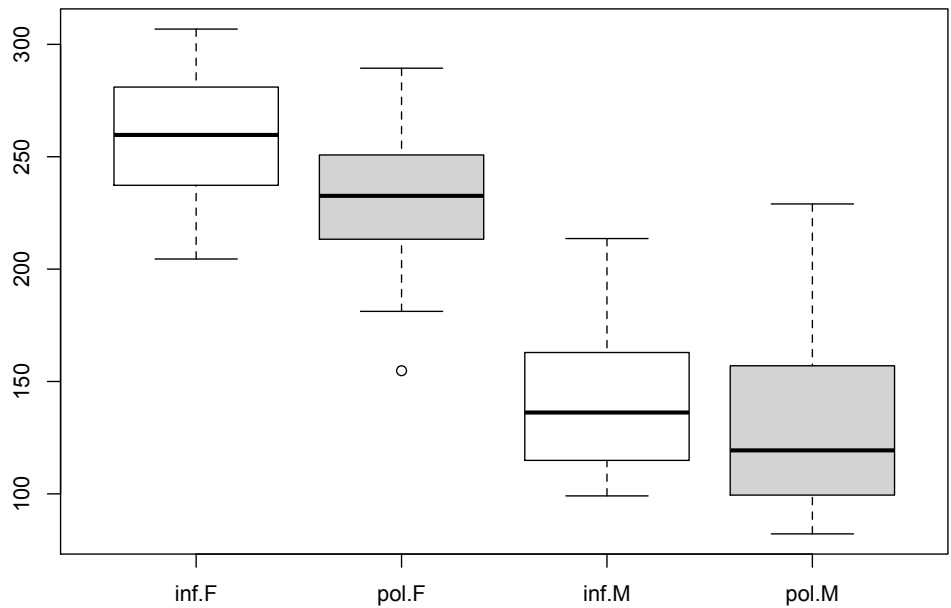

What do we see? In both cases, the median line (the thick line in the middle of the boxplot) is lower for the polite than for the informal condition. However, there may be a bit more overlap between the two politeness categories for males than for females.

Let's start with constructing our model!

Type in the command below ...

```
lmer(frequency ~ attitude, data=politeness)
```

... and you will retrieve an error that should look like this:

```
Error in lmerFactorList(formula, fr, rmInt = FALSE, drop = FALSE) :
  No random effects terms specified in formula
```

This is because the model *needs* a random effect (after all, "mixing" fixed and random effects is the point of mixed models). We just specified a single fixed effect, attitude, and that was not enough.

So, let's add random intercepts for subjects and items (remember that items are called "scenarios" here):

```
politeness.model = lmer(frequency ~ attitude +
    (1|subject) + (1|scenario), data=politeness)
```

The last command created a model that used the fixed effect "attitude" (polite vs. informal) to predict voice pitch, controlling for by-subject and by-item variability. We saved this model in the object `politeness.model`. To see what the result is, simply type in `politeness.model` to print the output (in contrast to `lm()` you don't need to use `summary()` to get this output).

This is the full output:

```
Linear mixed model fit by REML
Formula: frequency ~ attitude + (1 | subject) + (1 | scenario)
   Data: politeness
   AIC   BIC logLik deviance REMLdev
 803.5 815.5 -396.7   807.1   793.5
Random effects:
 Groups   Name        Variance Std.Dev.
 scenario (Intercept)  218.98  14.798
 subject  (Intercept) 4014.54  63.360
 Residual              646.02  25.417
Number of obs: 83, groups: scenario, 7; subject, 6

Fixed effects:
            Estimate Std. Error t value
(Intercept)  202.588     26.750   7.573
attitudepol  -19.695      5.585  -3.527

Correlation of Fixed Effects:
            (Intr)
attitudepol -0.103
```

Again, let's work through this: First, the output reminds you of the model that you fit. Then, there's some general summary statistics such as Akaike's Information Criterion, the log-Likelihood etc. We won't go into the meaning of these different values in this tutorial because these are conceptually a little bit more involved. Let's focus on the output for the random effects first:

```
Random effects:
 Groups   Name        Variance Std.Dev.
 scenario (Intercept)  218.98  14.798
 subject  (Intercept) 4014.54  63.360
 Residual              646.02  25.417
```

Have a look at the column standard deviation. This is a measure of the variability for each random effect that you added to the model. You can see that scenario ("item") has much less variability than subject. Based on our boxplots from above, where we saw more idiosyncratic differences between subjects than between items, this is to be expected. Then, you see "Residual" which stands for the variability that's not due to either scenario or subject. This is our "ε" again, the "random" deviations from the predicted values that are not due to subjects and items. Here, this reflects the fact that each and every utterance has some factors that affect pitch that are outside of the purview of our experiment.

The fixed effects output mirrors the coefficient table that we considered in tutorial 1 when we talked about the results of our linear model analysis.

```
Fixed effects:
            Estimate Std. Error t value
(Intercept)  202.588    26.750   7.573
attitudepol  -19.695     5.585  -3.527
```

The coefficient "attitudepol" is the slope for the categorical effect of politeness. Minus 19.695 means that to go from "informal" to "polite", you have to go down -19.695 Hz. In other words: pitch is lower in polite speech than in informal speech, by about 20 Hz. Then, there's a standard error associated with this slope, and a t-value, which is simply the estimate (20 Hz) divided by the standard error (check this by performing the calculation by hand).

Note that the `lmer()` function (just like the `lm()` function in tutorial 1) took whatever comes first in the alphabet to be the reference level. "inf" comes before "pol", so the slope represents the change from "inf" to "pol". If the reference category would be "pol" rather than "inf", the only thing that would change would be that the sign of the coefficient 19.695 would be positive. Standard errors, significance etc. would remain the same.

Now, let's consider the intercept. In tutorial 1, we already talked about the fact that oftentimes, model intercepts are not particularly meaningful. But this intercept is especially weird. It's 202.588 Hz … where does that value come from?

If you look back at the boxplot that we constructed earlier, you can see that the value 202.588 Hz seems to fall halfway between males and females – and this is indeed what this intercept represents. It's the average of our data for the informal condition.

As we didn't inform our model that there's two sexes in our dataset, the intercept is particularly off, in between the voice pitch of males and females. This is just like the classic example of a farm with a dozen hens and a dozen cows … where the mean legs of all farm animals considered together is three, not a particularly informative representation of what's going on at the farm.

Let's add gender as an additional fixed effect:

```
politeness.model = lmer(frequency ~ attitude +
    gender + (1|subject) +
    (1|scenario), data=politeness)
```

We overwrote our original model object `politeness.model` with this new model. Note that we added "gender" as a fixed effect because the relationship between sex and pitch is systematic and predictable (i.e., we expect females to have higher pitch). This is different from the random effects subject and item, where the relationship between these and pitch is much more unpredictable and "random". We'll talk more about the distinction between fixed and random effects later.

Let's print the model output again. Let's have a look at the residuals first:

```
Random effects:
 Groups   Name        Variance Std.Dev.
 scenario (Intercept) 219.45   14.814
 subject  (Intercept) 615.57   24.811
 Residual             645.90   25.414
```

Note that compared to our earlier model without the fixed effect gender, the variation that's associated with the random effect "subject" dropped considerably. This is because the variation that's due to gender was confounded with the variation that's due to subject. The model didn't know about males and females, and so it's predictions were relatively more off, creating relatively larger residuals.

Let's look at the coefficient table now:

```
Fixed effects:
            Estimate Std. Error t value
(Intercept)  256.846     16.114  15.940
attitudepol  -19.721      5.584  -3.532
genderM     -108.516     21.010  -5.165
```

We see that males and females differ by about 109 Hz. And the intercept is now much higher (256.846 Hz), as it now represents the female category (for the informal condition). The coefficient for the effect of attitude didn't change much.

**Statistical significance**

So far, we haven't talked about significance yet. But, if you want to publish this, you'll most likely need to report some kind of p-value. Unfortunately, p-values for mixed models aren't as straightforward as they are for the linear model. There are multiple approaches, and there's a discussion surrounding these, with sometimes wildly differing opinions about which approach is the best. Here, I focus on the Likelihood Ratio Test as a means to attain p-values.

Likelihood is the probability of seeing the data you collected given your model. The logic of the likelihood ratio test is to compare the likelihood of two models with each other. First, the model *without* the factor that you're interested in (the null model), then the model *with* the factor that you're interested in. Maybe an analogy helps you to wrap your head around this: Say, you're a hiker, and you carry a bunch of different things with you (e.g., a gallon of water, a flashlight). To know whether each item affects your hiking speed, you need to get rid of it. So, you get rid of the flashlight and run without it. Your hiking speed is not affected much. Then, you get rid of the gallon of water, and you realize that your hiking speed is affected a lot. You would conclude that carrying a gallon of water with you significantly affects your hiking speed whereas carrying a flashlight does not. Expressed in formula, you would want to compare the following two "models" (think "hikes") with each other:

>   mdl1 = hiking speed ~ gallon of water + flashlight
>   mdl2 = hiking speed ~ flashlight

If there is a significant difference between "mdl2" and "mdl1", then you know that the gallon of water matters. To assess the effect of the flashlight, you would have to do a similar comparison:

>   mdl1 = hiking speed ~ gallon of water + flashlight
>   mdl2 = hiking speed ~ gallon of water

In both cases, we compared a full model (with the fixed effects in question) against a reduced model without the effects in question. In each case, we conclude that a fixed effect is significant if the difference between the likelihood of these two models is significant.

Here's how you would do this in R. First, you need to construct the null model:

```
politeness.null = lmer(frequency ~ gender +
        (1|subject) + (1|scenario), data=politeness,
        REML=FALSE)
```

Note one additional technical detail. I just added the argument `REML=FALSE`. Don't worry about it too much – but in case you're interested, this changes some internal stuff (in particular, the likelihood estimator), and it is necessary to do this when you compare models using the likelihood ratio test (Pinheiro & Bates, 2000; Bolker et al., 2009).

Then, we re-do the full model above, this time also with `REML=FALSE`:

```
politeness.model = lmer(frequency ~ attitude +
        gender +
        (1|subject) + (1|scenario), data=politeness,
        REML=FALSE)
```

Now you have two models to compare with each other – one with the effect in question, one without the effect in question. We perform the likelihood ratio test using the `anova()` function:

```
anova(politeness.null,politeness.model)
```

This is the resulting output:

```
Data: politeness
Models:
politeness.null: frequency ~ gender + (1 | subject) + (1 | scenario)
politeness.model: frequency ~ attitude + gender + (1 | subject) + (1 | scenario)
                 Df    AIC    BIC  logLik  Chisq Chi Df Pr(>Chisq)
politeness.null   5 816.72 828.81 -403.36
politeness.model  6 807.10 821.61 -397.55 11.618      1  0.0006532 ***
```

You're being reminded of the formula of the two models that you're comparing. Then, you find a Chi-Square value, the associated degrees of freedom and the p-value[4].

---

[4] You might wonder why we're doing a Chi-Square test here. There's a lot of technical detail here, but the main thing is that there's a theorem, called Wilk's Theorem, which states that negative two times the log likelihood ratio of two models approaches a Chi-Square distribution with degrees of freedom of the number of parameters that differ between the models (in this case, only "attitude"). So, somebody has done a proof of this and you're good to go! Do note, also, that some people don't like "straight-jacketing" likelihood into the classical null-hypothesis significance testing framework that we're following here, and so they would disagree with the interpretation of likelihood the way we used it in the likelihood ratio test.

You would report this result the following way:

> "… politeness affected pitch ($\chi^2(1)$=11.62, p=0.00065), lowering it by about 19.7 Hz ± 5.6 (standard errors) …"

If you're used to t-tests, ANOVAs and linear model stuff, then this likelihood-based approach might seem weird to you. Rather than getting a p-value straightforwardly from your model, you get a p-value from a comparison of two models. To help you get used to the logic, remember the hiker and the analogy of putting one piece of luggage away to estimate that piece's effect on hiking speed.

Note that we kept the predictor "gender" in the model. The only change between the full model and the null model that we compared in the likelihood ratio test was the factor of interest, politeness. In this particular test, you can think of "gender" as a control variable and of "attitude" as your test variable.

We could have also compared the following two models:

full model:      frequency ~ attitude + gender
reduced model:   frequency ~ 1

"mdl.null" in this case is an intercept only model, where you just estimate the mean of the data. You could compare this to "mdl.full", which has two more effects, "attitude" and "gender". If this difference became significant, you would know that "mdl.full" and "mdl.null" are significantly different from each other – but you would not know whether this difference is due to "attitude" or due to "gender". Coming back to the hiker analogy, it is as if you dropped both the gallon of water and the flashlight and then you realized that your hiking speed changed, but you wouldn't be able to determine conclusively which one of the two pieces of luggage was the crucial one.

A final thing regarding likelihood ratio tests: What happens if you have an interaction? We didn't talk much about interactions yet, but say, you predicted "attitude" to have an effect on pitch that is somehow modulated through "gender". For example, it could be that speaking politely versus informally has the opposite effect for men and women. Or it could be that women show a difference and men don't (or vice versa). If you have such an inter-dependence between two factors (called an interaction), you can test it the following way:

full model:      frequency ~ attitude*gender
reduced model:   frequency ~ attitude + gender

In R, interactions between two factors are specified with a "*" rather than a "+".

If you compare the above models in a likelihood ratio test using the `anova()` function, then you would get a p-value that gives you the significance of the interaction. If this comparison is significant, you know that attitude and gender are significantly inter-dependent on each other. If this is comparison is not significant, there is no significant inter-dependence.

It might be a good idea to try out different likelihood comparisons with the data provided above, say "attitude*gender" versus "attitude + gender" versus simply "1" (the intercept only model). Remember to always put `REML=FALSE` when creating your model.

**Super-crucial: Random slopes versus random intercepts**

We're not done yet. One of the coolest things about mixed models is coming up now, so hang on!!

Let's have a look at the coefficients of the model by subject and by item:

```
coef(politeness.model)
```

Here is the output:

```
$scenario
  (Intercept) attitudepol   genderM
1    243.3398   -19.72111 -108.5164
2    263.4292   -19.72111 -108.5164
3    268.2541   -19.72111 -108.5164
4    277.4757   -19.72111 -108.5164
5    254.9102   -19.72111 -108.5164
6    244.6724   -19.72111 -108.5164
7    245.8426   -19.72111 -108.5164

$subject
   (Intercept) attitudepol   genderM
F1    242.9367   -19.72111 -108.5164
F2    267.2668   -19.72111 -108.5164
F3    260.3353   -19.72111 -108.5164
M3    285.2322   -19.72111 -108.5164
M4    262.2255   -19.72111 -108.5164
M7    223.0811   -19.72111 -108.5164
```

You see that each scenario and each subject is assigned a different intercept. That's what we would expect, given that we've told the model with "(1|subject)" and "(1|scenario)" to take by-subject and by-item variability into account.

But not also that the fixed effects (attitude and gender) are all the same for all subjects and items. Our model is what is called a ***random intercept model***. In this model, we account for baseline-differences in pitch, but we assume that whatever the effect of politeness is, it's going to be the same for all subjects and items.

But is that a valid assumption? In fact, often times it's not – it is quite expected that some items would elicit more or less politeness. That is, the effect of politeness might be different for different items. Likewise, the effect of politeness might be different for different subjects. For example, it might be expected that some people are more polite, others less. So, what we need is a ***random slope*** model, where subjects and items are not only allowed to have differing intercepts, but where they are also allowed to have different slopes for the effect of politeness. This is how we would do this in R:

```
politeness.model = lmer(frequency ~ attitude +
     gender + (1+attitude|subject) +
     (1+attitude|scenario),
     data=politeness,
     REML=FALSE)
```

Note that the only thing that we changed is the random effects, which now look a little more complicated. The notation "(1+attitude|subject)" means that you tell the model to expect differing baseline-levels of frequency (the intercept, represented by 1) as well as differing responses to the main factor in question, which is "attitude" in this case. You then do the same for items.

Have a look at the coefficients of this updated model by typing in the following:

```
coef(politeness.model)
```

Here's a reprint of the output that I got:

```
$scenario
  (Intercept) attitudepol     genderM
1    244.4725   -19.00303 -111.1032
2    261.9439   -12.87457 -111.1032
3    270.9277   -23.46232 -111.1032
4    277.0645   -15.90576 -111.1032
5    255.8264   -18.72596 -111.1032
6    247.0404   -22.37927 -111.1032
7    249.7023   -25.93020 -111.1032

$subject
   (Intercept) attitudepol     genderM
F1    243.2783   -20.49943 -111.1032
F2    267.1184   -19.30435 -111.1032
F3    260.2852   -19.64689 -111.1032
M3    287.1039   -18.30249 -111.1032
M4    264.6681   -19.42718 -111.1032
M7    226.3843   -21.34632 -111.1032
```

Now, the column with the by-subject and by-item coefficients for the effect of politeness ("attitudepol") is different for each subject and item. Note, however, that it's always negative and that many of the values are quite similar to each other. This means that despite individual variation, there is also consistency in how politeness affects the voice: for all of our speakers, the voice tends to go down when speaking politely, but for some people it goes down slightly more so than for others.

Have a look at the column for gender. Here, the coefficients do no change. That is because we didn't specify random slopes for the by-subject or by-item effect of gender.

O.k., let's try to obtain a p-value. We keep our model from above (`politeness.model`) and compare it to a new null model in a likelihood ratio test. Let's construct the null model first:

```
politeness.null = lmer(frequency ~ gender +
    (1+attitude|subject)   +   (1+attitude|scenario),
    data=politeness, REML=FALSE)
```

Note that the null model needs to have the same random effects structure. So, if your full model is a random slope model, your null model also needs to be a random slope model.

Let's now do the likelihood ratio test:

```
anova(politeness.null,politeness.model)
```

This is, again, significant.

There are a few important things to say here: You might ask yourself "Which random slopes should I specify?" … or even "Are random slopes necessary at all?"

A lot of people construct random intercept-only models but conceptually, it makes a hella sense to include random slopes most of the time. After all, you can almost always expect that people differ with how they react to an experimental manipulation! And likewise, you can almost always expect that the effect of an experimental manipulation is not going to be the same for all items.

Moreover, researchers in ecology (Schielzeth & Forstmeier, 2009), psycholinguistics (Barr, Levy, Scheepers, & Tilly, 2013) and other fields have shown via simulations that mixed models without random slopes are anti-conservative or, in other words, they have a relatively high Type I error rate (they tend to find a lot of significant results which are actually due to chance).

Barr et al. (2013) recommend that you should "keep it maximal" with respect to your random effects structure, at least for controlled experiments. This means that you include all random slopes that are justified by your experimental design … and you do this for all fixed effects that are important for the overall interpretation of your study.

In the model above, our whole study crucially rested on stating something about politeness. We were not interested in gender differences, but they are well worth controlling for. This is why we had random slopes for the effect of attitude (by subjects and item) but not gender. In other words, we only modeled by-subject and by-item variability in how politeness affects pitch.

## Assumptions

In tutorial 1, we talked a lot about the many different assumptions of the linear model. The good news is: Everything that we discussed in the context of the linear model applies straightforwardly to mixed models. So, you also have to worry about collinearity and influential data points. And you have to worry about homoscedasticity (and potentially about lack of normality). But you don't have to learn much new stuff. The way you check these assumptions in R is exactly the same as in the case of the linear model, say, by creating a residual plot, a histogram of the residuals or a Q-Q plot.

Independence, being the most important assumption, requires a special word: One of the main reasons we moved to mixed models rather than just working with linear models was to resolve non-independencies in our data. However, mixed models can still violate independence … if you're missing important fixed or random effects. So, for example, if we analyzed our data with a model that didn't include the random effect "subject", then our model would not "know" that there are multiple responses per subject. This amounts to a violation of the independence assumption. So choose your fixed effects and random effects carefully, and always try to resolve non-independencies.

Then, a word on influential data points. You will find that the function `dfbeta()` that we used in the context of linear models doesn't work for mixed models. If you worry about influential points, you can check out the package *influence.ME* (Nieuwenhuis, te Grotenhuis, & Pelzer, 2012), or you can program a for loop that does the leave-one-out diagnostics by hand. The following code gives you an outline of the general structure of how you might want to do this (you can check my "doodling" tutorials on loops and programming structures in R to get a better grasp of this):

```
all.res=numeric(nrow(mydataframe))
for(i in 1:nrow(mydataframe)){
    myfullmodel=lmer(response~predictor+
        (1+predictor|randomeffect),POP[-i,])
    all.res[i]=fixef(myfullmodel)[some number]
}
```
[5]

Go ahead and play with checking the assumptions. You can go back to tutorial 1 and apply the code in there to the new mixed model objects in this tutorial.

**A final note on random versus fixed effects**

I have evaded a precise definition of the difference between fixed and random effects. I deliberately did this because I wanted you to get some experience with linear mixed effects models in R before we finally take a step back and sharpen our concepts.

---

[5] The basic idea of this code snippet is this: Pre-define a vector that has as many elements as you have rows in your dataset. Then, cycle through each row. For each iteration, make a new mixed model *without that row* (this is achieved by `POP[-i,]`). Then, the function `fixef()` extracts whatever coefficient interests you.
You will need to adapt this code to your analysis. Besides the names of your data frame and your variables, you need to run `fixef()` on your model once so you know which position the relevant coefficient is. In our case, I would put a "2" in there because the effect of "attitudepol" appears second in the list of coefficients. "1" would give me the intercept, always the first coefficient mentioned in the coefficient table.

So, a random effect is generally something that can be expected to have a non-systematic, idiosyncratic, unpredictable, or "random" influence on your data. In experiments, that's often "subject" and "item", and you generally want to generalize over the idiosyncrasies of individual subjects and items.

Fixed effects on the other hand are expected to have a systematic and predictable influence on your data.

But there's more to it. One definition of fixed effects says that fixed effects "exhaust the population of interest", or they exhaust "the levels of a factor". Think back of sex. There's only "male" or "female" for the variable "gender" in our study, so these are the only two levels of this factor. Our experiment includes both categories and thus exhausts the category sex. With our factor "politeness" it's a bit trickier. You could imagine that there are more politeness levels than just the two that we tested. But in the context of our experiment, we *operationally defined* politeness as the difference between these two categories – and because we tested both, we fully "exhaust" the factor politeness (as defined by us).

In contrast, random effects generally sample from the population of interest. That means that they are far away from "exhausting the population" … because there's usually many many more subjects or items that you could have tested. The levels of the factor in your experiment is a tiny subset of the levels "out there" in the world.

## The write-up

A lot of tutorials don't cover how to write up your results. And that's a pity, because this is a crucial part of your study!!!

The most important thing: You need to describe the model to such an extent that people can **reproduce the analysis**. So, a useful heuristic for writing up your results is to ask yourself the question "Would I be able to re-create the analysis given the information that I provided?" If the answer is "yes" your write-up is good.

In particular, this means that you specify all fixed effects and all random effects, and you should also mention whether you have random intercepts or random slopes.

For reporting individual results, you can stick to my example with the likelihood ratio test above. Remember that it's always important to report the actual coefficients/estimates and not just whether an effect is significant. You should also mention standard errors.

Another important thing is to give enough credit to the people who put so much of their free time into making *lme4* and R work so efficiently. So let's cite them! It's also a good idea to cite exactly the version that you used for your analysis. You can find out your version and who to cite by typing in…

```
citation()
```

… for your R-version … and …

```
citation("lme4")
```

… for the lme4 package.

Finally, it's important that you mention that you checked assumptions, and that the assumptions are satisfied.

So here's what I would have written for the analysis that we performed in this tutorial:

> "We used R (R Core Team, 2012) and *lme4* (Bates, Maechler & Bolker, 2012) to perform a linear mixed effects analysis of the relationship between pitch and politeness. As fixed effects, we entered politeness and gender (without interaction term) into the model. As random effects, we had intercepts for subjects and items, as well as by-subject and by-item random slopes for the effect of politeness. Visual inspection of residual plots did not reveal any obvious deviations from homoscedasticity or normality. P-values were obtained by likelihood ratio tests of the full model with the effect in question against the model without the effect in question."

Yay, we're done!! I hope this tutorial was of help to you.